\UseRawInputEncoding
\documentclass[letterpaper, 10 pt, conference]{ieeeconf}
\IEEEoverridecommandlockouts
\usepackage{cite}
\usepackage{amsmath,amssymb,amsfonts}
\usepackage{algorithmic}
\usepackage{graphicx}
\usepackage{textcomp}
\usepackage{xcolor}
\usepackage{hyperref}

\def\BibTeX{{\rm B\kern-.05em{\sc i\kern-.025em b}\kern-.08em
    T\kern-.1667em\lower.7ex\hbox{E}\kern-.125emX}}

\usepackage{setspace}
\usepackage{caption}
\usepackage{bibentry}
\usepackage{color}
\usepackage{tabularray}
\definecolor{MineShaft}{rgb}{0.2,0.2,0.2}
\usepackage{booktabs}
\usepackage{booktabs, multirow} 
\usepackage{soul}
\newcommand\T{\rule{0pt}{2ex}}       
\newcommand\B{\rule[-1.2ex]{0pt}{0pt}} 
\usepackage{tikz}

\title{\LARGE \bf
Advancements in 3D Lane Detection Using LiDAR Point Clouds: From Data Collection to Model Development
}

\author{Runkai Zhao$^{1\ddagger}$, 
Yuwen Heng$^{2}$, 
Heng Wang$^{1}$,
Yuanda Gao$^{2}$, 
Shilei Liu$^{2}$,   
Changhao Yao$^{3\ddagger}$, \\
Jiawen Chen$^{2}$ 
and Weidong Cai$^{1}$
\thanks{$1$ School of Computer Science, University of Sydney,
\texttt{\{rzha9419,\newline hwan9147\}@uni.sydney.edu.au, tom.cai@sydney.edu.au}}%
\thanks{$2$ Baidu ACG, 
\texttt{\{hengyuwen,liushilei,\newline chenjiawen\}@baidu.com, yuanda.gao94@gmail.com}}
\thanks{$3$ School of Electronic Information and Electrical Engineering, Shanghai Jiao Tong University, 
\texttt{\{lio\_n1r\}@sjtu.edu.cn}}
\thanks{$\ddagger$ Work done during an internship at Baidu ACG}
}

\begin{document}

\maketitle
\thispagestyle{empty}
\pagestyle{empty}

\begin{abstract}
Advanced Driver-Assistance Systems (ADAS) have successfully integrated learning-based techniques into vehicle perception and decision-making. However, their application in 3D lane detection for effective driving environment perception is hindered by the lack of comprehensive LiDAR datasets. The sparse nature of LiDAR point cloud data prevents an efficient manual annotation process. To solve this problem, we present \textit{LiSV-3DLane}, a large-scale 3D lane dataset that comprises 20k frames of surround-view LiDAR point clouds with enriched semantic annotation. Unlike existing datasets confined to a frontal perspective, \textit{LiSV-3DLane} provides a full 360-degree spatial panorama around the ego vehicle, capturing complex lane patterns in both urban and highway environments. We leverage the geometric traits of lane lines and the intrinsic spatial attributes of LiDAR data to design a simple yet effective automatic annotation pipeline for generating finer lane labels. To propel future research, we propose a novel LiDAR-based 3D lane detection model, \textit{LiLaDet}, incorporating the spatial geometry learning of the LiDAR point cloud into Bird's Eye View (BEV) based lane identification. Experimental results indicate that \textit{LiLaDet} outperforms existing camera- and LiDAR-based approaches in the 3D lane detection task on the K-Lane dataset and our \textit{LiSV-3DLane}. The project code will be available at \url{https://github.com/RunkaiZhao/LiLaDet}. 
\end{abstract}

\section{INTRODUCTION}
\begin{figure}[t]
\centering
\includegraphics[width=\linewidth]{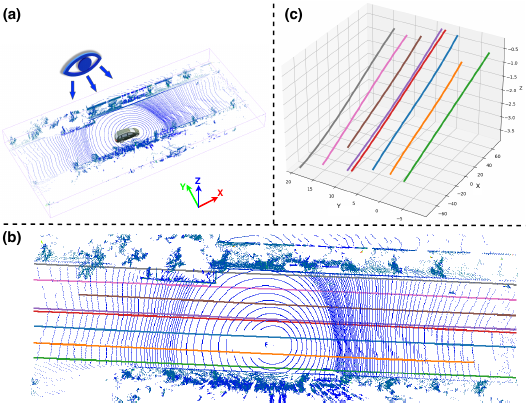}
\caption{
Our work aims at extracting 3D lane lines from a \textit{surround-view LiDAR point cloud} \textbf{(a)}. These detected lanes are visualized in BEV \textbf{(b)} and 3D coordinate system \textbf{(c)}.
}
\label{Fig: problem_statement}
\vspace{-2em}
\end{figure}

3D lane detection, which aims at providing accurate localization of lane lines in the real-world 3D coordinate system, offers spatial awareness for autonomous navigation and collision avoidance. Detecting lane lines in a surround view could provide a comprehensive understanding of traffic scenarios. LiDAR systems typically offer a 360-degree view of the environment, capturing lane information from all directions as shown in Fig. \ref{Fig: problem_statement}(a) and (b). However, the existing LiDAR-based 3D lane dataset \cite{paek2022k} only focuses on the frontal view due to the high cost of LiDAR data collection and labor-intensive annotation process caused by the sparse nature of point clouds \cite{bai2022build,kini20233dmodt}. 
To mitigate these challenges, this paper introduces a dedicated procedure for constructing a comprehensive LiDAR-based surround-view 3D Lane dataset from scratch.

\begin{figure}[t]
\centering
\includegraphics[width=\linewidth]{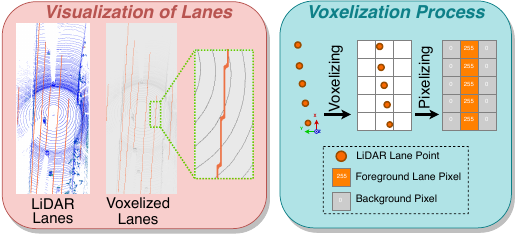}
\caption{ Current LiDAR-based 3D lane detection methods project LiDAR points into a BEV grid map. The voxelization process leads to the loss of spatial details due to the low resolution of discrete grid cells.
}
\label{Fig: motivation}
\vspace{-2em}
\end{figure}

Existing camera-based 3D lane detection methods are ill-posed as 2D images cannot be converted to 3D representations without depth information. \cite{garnett20193d,efrat20203d,guo2020gen,chen2022persformer,wang2023bev,huang2023anchor3dlane} assume that the ground is flat and assign zero height to all lanes. This planar assumption cannot be generalized to real-world driving conditions such as non-linear terrains with slopes and bumps. On the other hand, LiDAR-based 3D lane detection methods \cite{paek2022k,paek2022row} project LiDAR points into a Bird's Eye View (BEV) grid image and identify line lanes by semantic segmentation, which can generally display lane shapes but fail to capture accurate 3D information due to the reduced spatial details of the voxelization process as illustrated in Fig. \ref{Fig: motivation}.  

In this paper, we present a \textbf{\textit{Li}}DAR-based \textbf{\textit{S}}urround-\textbf{\textit{V}}iew \textbf{\textit{3D}} \textbf{\textit{Lane}} dataset, namely \textbf{\textit{LiSV-3DLane}}, collecting 20,025 frames of point clouds with 3D lanes manually annotated. Compared to existing datasets, ours contains omnidirectional LiDAR points. Since the manual annotation only captures sparse geometry information of lane lines, we also design an automatic annotation pipeline to generate finer lane labels specifically for dense prediction tasks. This pipeline can be used for a single-frame point cloud by harnessing lane spatial geometries and point cloud attributes (intensity and coplanarity). Lastly, we propose a novel \textbf{\textit{Li}}DAR-based 3D \textbf{\textit{La}}ne \textbf{\textit{Det}}ection framework, dubbed as \textbf{\textit{LiLaDet}}, which is designed for lane semantic feature and geometry learning. Given a point cloud as input, our model first identifies the lane segments from the projected BEV space to generate 3D lane point proposals \textit{(BEV Pathway)}. Then, to complement the spatial detail loss caused by voxelization, we design a \textit{Spatial Pathway} to refine the lane point proposals with geometric regression and confidence prediction, which accurately restores the 3D positions.

Our main contributions include: \textbf{i}) we introduce a LiDAR-based surround-view lane dataset, \textit{LiSV-3DLane} encompassing 20k frames that capture diverse and sophisticated urban and highway scenes; \textbf{ii}) we propose an automatic lane annotation pipeline to enrich the acquired manual annotation and generate finer lane annotation, leveraging the inherent lane streamline geometry and intrinsic attributes of spatial points; \textbf{iii}) we design a novel LiDAR-based framework, \textit{LiLaDet}, to facilitate the identification of lane markings with point cloud data, integrating both BEV and 3D spatial perspectives to achieve accurate lane identification and localization.

\section{RELATED WORKS}
3D lane detection was initially developed with camera-based methods among which the pioneering work in the camera-based 3D lane detection domain is 3D-LaneNet \cite{garnett20193d}. It first encodes hierarchical image features and then projects these features onto a virtual top-view plane by using camera parameters. Gen-LaneNet \cite{guo2020gen} introduces an expandable two-stage framework, where it first segments lane pixels from the 2D frontal image and then applies convolutional layers to learn the geometric transformation required to restore the 3D lanes. Unlike previous methods, \cite{chen2022persformer, huang2023anchor3dlane, wang2023bev} add strong spatial and structural priors into lane image feature extraction. Anchor3DLane \cite{huang2023anchor3dlane} employs frontal image features to directly regress lane anchors that are defined in 3D space. However, the success of monocular 3D lane detection heavily relies on the flat ground assumption to restore 3D information, which prevents them from estimating the depth precisely.

Compared to cameras, LiDAR sensors can offer surround-view 3D perception in varying lighting and weather conditions \cite{hahner2021fog,hahner2022lidar}. However, extracting lane lines from LiDAR point data is challenging due to the sparse nature of the point cloud. To handle this problem, recent works \cite{Liu2023Learning,guan2023flexible} explore the fusion of multiple frames captured at different timestamps and viewpoints, projecting points onto the BEV plane to detect lane marks with segmentation-based methods. However, in practical applications, multi-sensor synchronization is challenging to achieve with appropriate sequential data augmentation due to the high-expense collaborative calibration. \cite{paek2022row} proposes a two-stage LiDAR lane detection network, incorporating a row-wise BEV lane feature learning and a local lane correlation refinement. As the first LiDAR-based 3D lane dataset, K-Lane \cite{paek2022k} only provides planar lane annotations on a downscaled BEV space and the spatial loss of voxelization is inevitable as shown in Fig. \ref{Fig: motivation}. In conclusion, the realm of LiDAR-based 3D lane detection remains underexplored for holistic driving scene understanding in both dataset and model development. To boost the development, we bring in the first large-scale LiDAR-based surround-view 3D lane dataset (\textit{LiSV-3DLane}). An automatic lane annotation pipeline is designed to generate finer lane annotation with richer semantic details. To better learn the 3D lane semantic and geometric features, we develop a novel LiDAR-based 3D lane detection framework \textit{LiLaDet} which is robust to various driving scenarios. 

\section{LiSV-3DLANE DATASET}
\subsection{Dataset Introduction}
\noindent\textbf{Raw LiDAR Data.} \textit{LiSV-3DLane} is a comprehensive LiDAR point cloud dataset that focuses particularly on surround-view 3D lane data. It comprises 20,025 frames from 1,003 unique driving sequences. The dataset contains different daytime periods including morning, afternoon, dusk, and night, and various lighting conditions including sunny, cloudy, and rainy. Besides normal road conditions, it also captures challenging driving scenarios for urban and highway areas such as crowded traffic zones and under-construction roads. These diversified conditions lead to varying degrees of lane occlusions which could enhance the robustness and generalizability in the training of learning-based lane detection models. 

\noindent\textbf{Sensor Suite.}  \textit{LiSV-3DLane} is collected using a Velodyne VLS-128 LiDAR sensor with 128 channels and 0.1$\sim$0.4 degree horizontal angular resolution, and seven cameras with 3840$\times$2160 image resolutions. These sensors are finely calibrated and synchronized to ensure high data quality.

\begin{figure}[t]
\centering
\includegraphics[width=\linewidth]{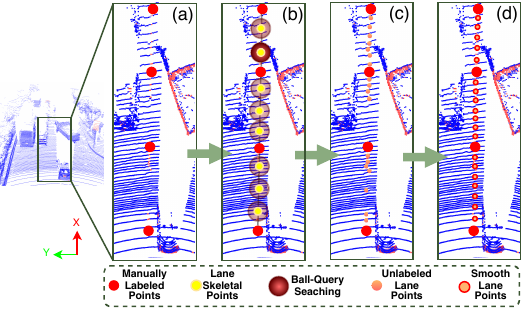}
\caption{
   \textbf{Automatic Lane Annotation Pipeline for generating finer lane annotation.} \textbf{(a)} Raw manual lane annotation; \textbf{(b)} Lane skeleton representations and lane skeletal points equidistantly sampled along links; \textbf{(c)} Unlabeled lane points selected by ball-query searching; \textbf{(d)} Smooth lane points sampled from the interpolated cubic curve function.
}
\label{Fig: lane_skeleton}
\vspace{-2em}
\end{figure}

\noindent\textbf{Lane Manual Annotation.} The ground-truth lanes are annotated by qualified specialists following the acknowledged lane annotation standard \cite{yan2022once,chen2022persformer,paek2022k}. A lane line in 3D space is manually annotated as a set of points to demarcate drivable zones and is represented by $\{[x_{i}, y_{i}, z_{i}]\}^{N_{p}}_{i=1}$ where $N_{p}$ is the number of lane points. Although sparse point-wise annotation is amenable to manual labeling, it lacks density and continuity. Capturing only discrete locations along a lane, as shown in Fig. \ref{Fig: lane_skeleton}(a), the manually acquired annotation yields low-level geometry information, which constrains contextual understanding of a full driving environment (e.g., how a curve lane functions in relation to the overall traffic flow). To address this limitation, we propose an automatic annotation pipeline that creates detailed lane shapes and considers their inherent point cloud characteristics (intensity and coplanarity). This automatic pipeline can be applied to other LiDAR-based 3D lane datasets to produce dense lane points and provide deeper insights into lane features.

\subsection{Automatic Lane Annotation Pipeline}
The current techniques used to annotate lanes in LiDAR point clouds rely on merging multiple sequential frames through Simultaneous Localization And Mapping (SLAM) \cite{Liu2023Learning,guan2023flexible}. Yet, this method demands a high-expense time synchronization approach. Based on our observation, lane points can be identified with their spatial geometries and point cloud attributes. Specifically, we use the RANdom SAmple Consensus (RANSAC) validation step, lane skeletal abstraction, ball-query searching, and cubic curve interpolation to produce finer lane annotation as shown in Fig. \ref{Fig: lane_skeleton}.

\noindent\textbf{RANSAC Neighboring Ground Plane Fitting.} To assess the accuracy and reliability of manually annotated lane points, we introduce a validation step based on the RANSAC algorithm \cite{fischler1981random}. It can identify the nearest terrain surface plane corresponding to each manually labeled lane point. After calculating each point's perpendicular distance to the plane, a pre-defined distance threshold of 0.01 meters is adopted to determine if this lane point can be accepted. To bolster ground truth quality, erroneous lane points are projected onto the identified plane for height recalibration. The RANSAC algorithm is implemented by using the open-source machine-learning library \texttt{scikit-learn}.

\noindent\textbf{Lane Skeletonization.} Inspired by recent works \cite{gao2020vectornet,li2022hdmapnet} modeling driving scene components as connected polylines, we characterize lane geometric shape by linking lane points in the positive $x$ direction and sampling skeletal points equidistantly along each link as shown in Fig. 3(b). 

\noindent\textbf{Ball-Query Lane Points Searching.} Assuming the lane skeletal point as a reference centroid, we employ ball query searching to identify local unlabeled lane points as shown in Fig. \ref{Fig: lane_skeleton}(b) and (c). Attributed to the reflectivity of lane paint material, LiDAR lane points typically exhibit higher intensity than ground points. However, other points on curbs or shrubs also have relatively high intensity and can be misclassified as lane points. To mitigate this confusion, we incorporate coplanarity as a criterion to filter incorrect lane points. 

\noindent\textbf{Cubic Curve Interpolation.} Lastly, the smooth lane points are sampled from an interpolated cubic curve function as shown in Fig. \ref{Fig: lane_skeleton}(d), which offers a complete lane shape with richer geometric details.

\begin{figure}
\centering
\includegraphics[width=\linewidth]{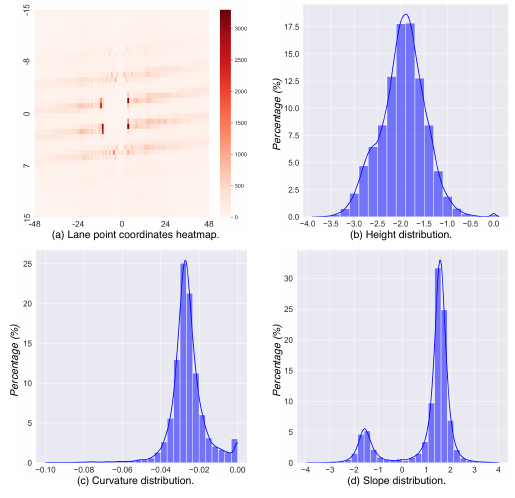}
\caption{
\textbf{Dataset Statistics Analysis.} We analyze the coordinates, height, curvature, and slope of lanes to illustrate the diversity of lane geometry in \textit{LiSV-3DLane}.
}
\label{Fig: dataset_statics}
\vspace{-2em}
\end{figure}

\begin{figure*}[t]
\centering
\includegraphics[width=0.95\textwidth]{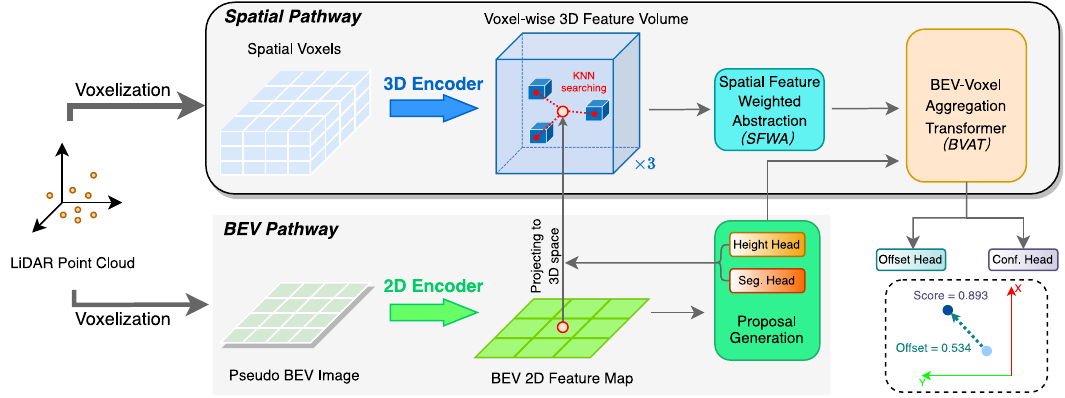}
\caption{
\textbf{Overview of our proposed \textit{LiLaDet} framework.} Given a LiDAR point cloud as input, our model first identifies the lane segments from the
projected BEV space to generate 3D lane point proposals
at the \textit{BEV Pathway}. Then, we design a \textit{Spatial Pathway} to refine the lane proposal points through geometric regression and confidence prediction.
}
\label{Fig: framework}
\vspace{-1.5em}
\end{figure*}

\subsection{Dataset Analysis} 
\label{subsection: dataset_analysis} In our \textit{LiSV-3DLane} dataset, lane points are annotated within the LiDAR coordinate system, so we focus on the analysis of their 3D positional attributes. Fig. \ref{Fig: dataset_statics}(a) visualizes 2D coordinates of all lane points using a heatmap. The horizontal axis is the $x$ coordinate, the vertical axis is the $y$ coordinate, and the frequency of lane point occurrences is indicated by color shade. From the visualization, it is apparent that the labeled lanes surround the ego vehicle, facilitating surround-view contextual learning for model development. Different from the existing LiDAR-based lane dataset \cite{paek2022k}, we provide lane labels with height values, enabling the model to predict realistic 3D lane coordinates. Since the LiDAR sensor is placed at the top of the ego vehicle, the height values of lanes are all negative and their distribution is illustrated in Fig. \ref{Fig: dataset_statics}(b). 
A set of 3D lane points is often interpolated as a cubic curve function to model their geometric shape \cite{guo2020gen}. Taking $x$-$y$ plane as an example, the curve function is expressed as $\small{y=ax^3+bx^2+cx+d}$ where the primary controllable parameter $a$ significantly influences lane curvature. Its value distribution is illustrated in Fig. \ref{Fig: dataset_statics}(c). It is worth noting that lanes exhibit diverse curvatures and they do not fit a uniform geometric shape. This variability hinders the lane geometry understanding. The features of lanes, including their coordinates throughout the whole scene and their varied shapes, can explain that the anchor-based method using 3D bounding boxes as in \cite{lang2019pointpillars} or line-like structures as in \cite{huang2023anchor3dlane} is not suitable for lane detection. Extensive pre-defined anchors of various shapes are required to cover each possible position, which is impractical in model convergence and operation time. Camera-based 3D lane detection adopts a flat ground assumption to collerate 2D frontal image to BEV space. As Fig. \ref{Fig: dataset_statics}(d) indicates, lanes in real-world scenarios rarely conform to a perfectly horizontal alignment, thereby underscoring the limitations of camera-based models for capturing reliable spatial geometry.

\section{LILADET FRAMEWORK}
In this study, we introduce \textit{LiLaDet}, a novel LiDAR-based 3D lane detection framework. As schematically illustrated in Fig. \ref{Fig: framework}, the framework consists of two pathways: the BEV and the spatial pathways, which operate on two voxelized representations of the LiDAR point cloud input: pseudo BEV grid image and spatial voxels, respectively. In the \textit{BEV Pathway}, we train a 2D encoder to extract BEV feature maps, leveraging the global receptive field of the attention mechanism to capture spatial dependencies across the scene. These feature maps subsequently yield a segmentation map and height prediction for lane proposal generation. In the \textit{Spatial Pathway}, a stack of 3D sparse convolutions with multiple scales encodes the input into hierarchical spatial features. Then, at the three deepest scales, features are separately searched using the \textit{k} Nearest Neighbour (\textit{k}-NN) algorithm to generate local spatial features around lane proposal points \cite{shi2020pv}. We introduce a Spatial Feature Weighted Abstraction \textit{(SFWA)} module for weighted aggregation of multi-scale local spatial features. Lastly, the BEV and spatial features are merged in a BEV-Voxel Aggregation Transformer \textit{(BVAT)} module which employs a cross-attention mechanism to decode features from different modalities when making final lane predictions. The details of our proposed framework are explained below.

\subsection{2D BEV Pathway}
In BEV segmentation methods \cite{bai2022build,paek2022k,liu2023bevfusion,peng2023bevsegformer}, traffic entities of arbitrary shapes are identified using a set of foreground pixels in the BEV space. Such a pixel-wise representation provides adaptability to the inherent lane geometry variances, which prevents lane detection from the human-crafted spatial and structural priors or anchors. Inspired by this, we formulate lane detection as a semantic segmentation problem. Concretely, we first generate a pseudo BEV grid image by pillarizing raw LiDAR points \cite{yang2018pixor}.
Analogous to 2D image processing, we employ a Vision Transformer (ViT) \cite{dosovitskiy2020image,fan2021multiscale} with a multi-scale receptive field to learn discriminative BEV feature maps $\small{F_{bev}\in \mathbb{R}^{L_{bev}\times W_{bev} \times C_{bev}}}$ where $L_{bev}$ and $W_{bev}$ denote the size of the BEV feature map and $C_{bev}$ denotes the feature channel dimension.

\noindent\textbf{Lane Point Proposal Generation.} We forward BEV feature maps $F_{bev}$ into a 2D convolutional segmentation head to partition foreground lane pixels from the grid scene. To obtain realistic 3D coordinates of lane lines, we additionally append a shared Multi-Layer Perceptron (MLP) head to predict the height of each positive lane pixel with residual learning. Given $xy$ coordinates in LiDAR space calculated from the segmentation map and $z$ coordinate of the height map, we can project the lane pixels of the 2D BEV plane back to 3D space, generating 3D lane proposal points. 

\subsection{3D Spatial Pathway}
Semantic segmentation can effectively capture lane instances from LiDAR points, but the generated lane proposal points in the \textit{BEV Pathway} are not accurately localized in 3D space. Since the original BEV grid image is encoded to obtain the low-resolution BEV feature maps, such coarse BEV maps cannot provide sufficient spatial details to restore accurate lane localization in the input scene \cite{ronneberger2015u,shi2020pv}. To mitigate this problem, we design the 3D \textit{Spatial Pathway} to complement the \textit{BEV Pathway} by directly using LiDAR point cloud. This explicit 3D data representation provides vital spatial geometric cues for further lane proposal refinement. We employ 3D voxel CNN with sparse convolutions \cite{yan2018second,shi2020pv} to efficiently encode a LiDAR point cloud into hierarchical spatial feature volumes. The raw LiDAR point cloud is first discretized into a set of 3D volumes. The 3D encoder stacks $3\times3\times3$ sparse convolutional layers to gradually process voxelized point cloud into 3D spatial feature volumes $F_{sp}^{l_{n}}$ at different scale levels $l_{n}$ where $n$ is the scale level order (i.e., $n$ = 1, 2, 3, 4). 

\noindent\textbf{Spatial Feature Weighted Abstraction \textit{(SFWA)}.} 
As the receptive field enlarges as the 3D voxel CNN network goes deeper, voxel-wise spatial features at different scale levels exert different influences on spatial contextual understanding. In open wild scenarios, lane detection is sensitive to fine-scale voxel-wise spatial features, such as the adjacent road surface features. Conversely, in complex traffic situations, spatial features at the coarse scale can better capture long-range spatial relationships with large objects such as vehicles or other traffic infrastructures. 

To capture the local voxel-wise spatial features surrounding each lane proposal point, a set abstraction module like \cite{qi2017pointnet,shi2020pv} is employed at each scale level. We utilize the \textit{k}-NN algorithm to retrieve \textit{k}-nearest voxel-wise features and concatenate them as $\small{\{\mathcal{F}^{l_{n}} \in \mathbb{R}^{\textit{k}\times C_{v}}: [f_{1}^{{l_{n}}};\cdots;f_{\textit{k}}^{{l_{n}}}]\}}$ where $C_{v}$ is the channel dimension of the voxel-wise spatial feature. Considering transform invariance in point cloud processing \cite{qi2017pointnet}, we use the max-pooling operation to generate a local abstract feature vector, denoted as $\small{\{S^{l_{n}}\in\mathbb{R}^{1\times C_{v}}: \mathcal{P}(\mathcal{F}^{l_{n}})\}}$ where $\mathcal{P}$ is the max-pooling operation. In our case, we only consider the spatial features at the $l_{2}$,  $l_{3}$, and $l_{4}$ scale levels. This abstract module is performed at each scale level, then all abstract features are concatenated as $\small{\{S\in \mathbb{R}^{3\times C_{v}}:[S^{l_{2}};S^{l_{3}};S^{l_{4}}]\}}$. Afterwards, three attention weights $\small{W\in\mathbb{R}^3}$ are predicted, emphasizing the scale level contributing the most in lane spatial positional learning as follows: $\small{W=\sigma(\mathcal{M}(S))}$ where $\mathcal{M}$ stands for a shared MLP layer and $\sigma$ denotes the softmax function, then these weights are separately multiplicated with the corresponding voxel-wise spatial feature $\mathcal{F}^{l_{n}}$. The weighted features from all scale levels are lastly concatenated and processed by another shared MLP to generate final spatial features for a lane proposal point: $\small{\{F_{sp} \in \mathcal{R}^{1\times C_{sp}}: \mathcal{M}(\mathcal{R}([\hat{\mathcal{F}^{l_{2}}};\hat{\mathcal{F}^{l_{3}}};\hat{\mathcal{F}^{l_{4}}}]))\}}$ where $\mathcal{R}$ denotes the reshape operation and $C_{sp}$ is the spatial feature channel dimension.

\noindent\textbf{BEV-Voxel Aggregation Transformer \textit{(BVAT)}.} The Cross-Attention (CA) mechanism is employed to fuse and calculate the correlation between multiple feature resources in vision tasks \cite{dosovitskiy2020image,liu2022petrv2,li2022bevformer,li2022deepfusion}, which can enrich a query $\mathbf{Q}$ vector with complementary information provided by a pair of key $\mathbf{K}$ and value $\mathbf{V}$ vectors. Specifically, this process can be defined as the following functions:
\begin{gather}
    \Bar{\mathbf{Q}}=\mathbf{Q}\mathbf{W_{q}}, \Bar{\mathbf{K}}=\mathbf{K}\mathbf{W_{k}},
    \Bar{\mathbf{V}}=\mathbf{V}\mathbf{W_{v}}\\    \text{CA}(\Bar{\mathbf{Q}},\Bar{\mathbf{K}},\Bar{\mathbf{V}})= softmax(\Bar{\mathbf{Q}}\Bar{\mathbf{K}}^{T}/\sqrt{D_{h}})\Bar{\mathbf{V}}
\label{eq:1}
\end{gather}
\noindent where $\mathbf{W_{\{q,k,v\}}}$ is linear transformations and $D_{h}$ denotes the hidden feature embedding dimension.
In our case, CA is applied across BEV grid image and spatial voxels. Suppose $N$ lane proposal points are generated from the \textit{BEV Pathway}, we have 2D BEV feature and 3D spatial feature with the size being $N\times C_{bev}$ and $N\times C_{sp}$ respectively. \textit{BVAT} module aims at fusing these two features from different data modalities. Concretely, we generate the module inputs as:
\begin{gather} 
\mathbf{Q}_{bev}, \mathbf{K}_{bev}, \mathbf{V}_{bev}= \text{LN}(F_{bev}), \text{LN}(F_{bev}), \text{LN}(F_{bev})\\
\mathbf{Q}_{sp}, \mathbf{K}_{sp}, \mathbf{V}_{sp}= \text{LN}(F_{sp}), \text{LN}(F_{sp}), \text{LN}(F_{sp}) 
\label{eq:2}
\end{gather}

\noindent where LN denotes Layer Normalization. The CA across two modalities is manipulated as:
\begin{equation}
\begin{split}
\mathbf{Z} = &\text{FFN}(\text{CA}(\mathbf{Q}_{bev}, \mathbf{K}_{sp}, \mathbf{V}_{sp})) \\
&+ \text{FFN}(\text{CA}(\mathbf{Q}_{sp}, \mathbf{K}_{bev},\mathbf{V}_{bev}))
\end{split}
\label{eq:3}
\end{equation}

\noindent where \text{FFN} denotes Feed Forward Network and $\mathbf{Z}$ denotes output feature. Then, the MLP-based offset head and confidence head are attached to predict the \textit{xy} coordinate offset and the confidence score of each lane point, respectively.

\subsection{Learning Objectives}
We use Binary Cross Entropy (BCE) loss to supervise the training of the segmentation head to partition pixel-wise lanes \cite{ronneberger2015u,paek2022k}. For the geometric regression training, we use smooth L1-Norm loss \cite{girshick2015fast} to supervise height prediction in the \textit{BEV Pathway} and \textit{xy} coordinate offset prediction in the \textit{Spatial Pathway} \cite{shi2020pv,yang2018pixor}. Notably, lane BEV segmentation output may yield false positive pixels (i.e. turning road markings) that have similar intensity as lane markings. To counteract this problem, a confidence head is introduced to geometrically select high-fidelity lane points based on predicted scores. We compute the Euclidean Distance between a lane proposal point and its nearest ground-truth point. If the distance is within a distance threshold $\tau$, this point is assigned as true, otherwise, it is false. The confidence prediction head is trained using BCE loss.

\section{EXPERIMENTS AND RESULTS}
\subsection{Datasets and Metrics} 
We split our \textit{LiSV-3DLane} dataset into 12,000/3,982/4,043 frames for training/validation/testing sets. We also conduct experiments on K-Lane\cite{paek2022k} to evaluate lane detection performance. We split a total of 15,382 frames into 7,687/3,848/3,847 frames for training/validation/testing sets, but the dataset only provides lane labels on the BEV plane without realistic height values. For effective 3D lane detection evaluation, given a labeled lane point in K-Lane, we assume the minimum height value of its neighboring points as the true height value. We employ standard 3D lane detection measures \cite{guo2020gen} using the bipartite matching method to match the predicted and ground-truth lanes for calculating precision, recall, and F1-score metrics. To calculate spatial similarity, we also use unilateral Chamfer Distance (CD), a common distance metric in point cloud processing tasks.

\subsection{Implementation Details}
The BEV and spatial pathways are trained separately using the AdamW optimizer with a learning rate of $2e^{-4}$. We use a small \textit{xy} resolution of 0.04 m to form a pseudo BEV input image at $\text{2400}\times\text{1000}$ resolution \cite{lang2019pointpillars,paek2022k}, and 0.32 m to form a downscaled lane ground-truth image. The \textit{BEV Pathway} is trained with a batch size of 4 for 24 epochs. When testing, we use a density-based spatial clustering method to classify lane instances. In the \textit{Spatial Pathway}, we voxelize the point cloud with a voxel size of [0.1, 0.1, 0.2], a maximum number of points per voxel of 32, and a maximum number of 12000. Given the lane point proposals, we search for \textit{k}=12 neighboring voxel features to capture local spatial information. The distance threshold $\tau$ is 0.5 meters. The \textit{Spatial Pathway} is trained with a batch size of 20 for 6 epochs. All experiments are conducted on NVIDIA GeForce RTX 3090s and PyTorch 1.11.0.

\begin{figure}[t]
\centering
\includegraphics[width=0.9\linewidth]{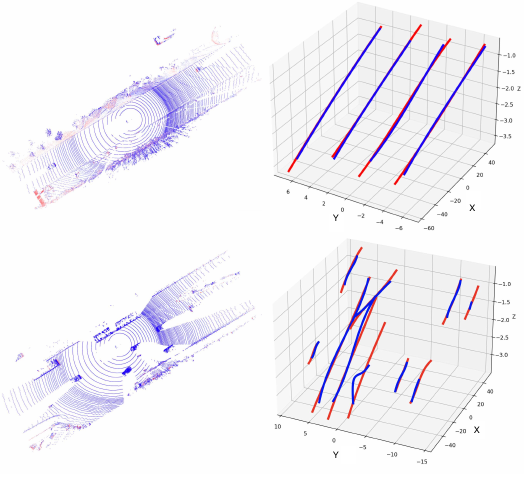}
\caption{
Qualitative Evaluation of \textit{LiLaDet} on \textit{LiSV-3DLane}. \textbf{Left}: LiDAR point cloud inputs; \textbf{Right}: The predicted and ground-truth lanes are shown in \textcolor{blue}{blue} and \textcolor{red}{red}, respectively.
}
\label{Fig: qualitative_eval}
\vspace{-3ex}
\end{figure}

\subsection{Results and Analysis} 
To investigate the applicability of our proposed framework, we evaluate lane detection models under frontal and omnidirectional views and report their performances in Tables \ref{table:2} and \ref{table:3}. However, camera-based 3D lane detection methods have a limited field of view and strict planar assumptions of bridging 2D and 3D space, which suffer in realistic application. Turning to LiDAR-based methods, they are extrinsic-free, and directly process real-world data, but the existing LiDAR-based methods only identify lanes in BEV space without height predictions. To make a comprehensive performance analysis in spatial measures, we re-implement LLDN-GFC by adding a height prediction head as indicated by $\dagger$. The comparisons show our proposed method outperforms other existing LiDAR-based methods under frontal and omnidirectional views. We also examine our model performance on the K-Lane dataset in Table \ref{table:4}. Compared with other LiDAR-based lane detection methods, with the fusion of BEV and 3D spatial information, our model can effectively detect and localize the lanes in the omnidirectional scene.

In the qualitative evaluation in Fig. \ref{Fig: qualitative_eval}, our model has robust detection performance for urban and highway scenarios. However, in complex driving scenarios containing more traffic components, the insufficient number of point clouds allocated to lane lines fails to fully reveal the necessary spatial geometric features, leading to suboptimal solutions.

\begin{table}[ht]
\centering
\caption{Experimental results on our \textit{LiSV-3DLane} test set for Frontal View.}
\resizebox{0.95\columnwidth}{!}{%
\begin{tabular}{@{}ccccccc@{}}
\toprule
$Method$ & $Modility$ & $Precision(\%)\uparrow$ & $Recall(\%)\uparrow$ & $F1(\%)\uparrow$    & $CD_{3D}($m$)\downarrow$ & $CD_{{\tiny BEV}}($m$)\downarrow$ \\ 
\midrule
3D-LaneNet\cite{garnett20193d}            & Image &51.41 &22.82 &31.61 &0.940 &0.940 \T\B \\
Gen-LaneNet\cite{guo2020gen}             & Image &54.15 &19.36 &28.53 &0.260 &0.251 \T\B \\
Anchor3DLane\cite{huang2023anchor3dlane} & Image &57.23      &33.40 &42.18      &0.312     &0.301   \T\B \\ 
LLDN-GFC\cite{paek2022k}              & LiDAR &65.79 &79.92 &72.17 &-     &0.232 \T\B \\ 
$^\dagger$LLDN-GFC\cite{paek2022k}    & LiDAR &\textbf{65.81} &79.94 &72.19 &0.235 &0.232 \T\B \\ 
RLLDN-LC\cite{paek2022row}              & LiDAR &62.20 &78.64 &69.46 &- &0.203 \T\B \\ 
LiLaDet (ours)        & LiDAR &65.78 &\textbf{85.16} &\textbf{74.23} &\textbf{0.158} &\textbf{0.158} \T\B \\
\bottomrule
\end{tabular}%
}
\label{table:2}
\end{table}

\begin{table}[ht]
\centering
\caption{Experimental results on our  \textit{LiSV-3DLane} test set for Surround View.}
\resizebox{0.95\columnwidth}{!}{%
\begin{tabular}{@{}ccccccc@{}}
\toprule
$Method$ & $Modility$ & $Precision(\%)\uparrow$ & $Recall(\%)\uparrow$ & $F1(\%)\uparrow$    & $CD_{3D}($m$)\downarrow$ & $CD_{BEV}($m$)\downarrow$ \\ 
\midrule
LLDN-GFC\cite{paek2022k}              & LiDAR &61.25 &78.18 &68.69 &-     &0.198 \T\B \\ 
$^\dagger$LLDN-GFC\cite{paek2022k}    & LiDAR &61.28 &78.25 &68.73 &0.195 &0.192 \T\B \\ 
RLLDN-LC\cite{paek2022row}              & LiDAR &58.44 &73.39 &65.07 &- &0.179 \T\B \\ 
LiLaDet (ours)        & LiDAR &\textbf{63.68} &\textbf{83.76} &\textbf{72.35} &\textbf{0.150} &\textbf{0.147} \T\B \\
\bottomrule
\end{tabular}%
}
\label{table:3}
\end{table}

\begin{table}[!htb]
\centering
\caption{Experimental results on K-Lane test set.}
\resizebox{0.95\columnwidth}{!}{%
\begin{tabular}{@{}ccccccc@{}}
\toprule
$Method$ & $Modility$ & $Precision(\%)\uparrow$ & $Recall(\%)\uparrow$ & $F1(\%)\uparrow$    & $CD_{3D}($m$)\downarrow$ & $CD_{BEV}($m$)\downarrow$ \\ \midrule
3D-LaneNet\cite{garnett20193d}            & Image &78.28 &27.81 &41.04 &0.674 &0.637 \T\B \\
Gen-LaneNet\cite{guo2020gen}            & Image &\textbf{82.37} &29.55 &43.49 &0.302 &0.294 \T\B \\
LLDN-GFC\cite{paek2022k}              & LiDAR &70.37 &85.59 &77.24 &- &0.230 \T\B \\ 
$^\dagger$LLDN-GFC\cite{paek2022k}    & LiDAR &70.37 &85.62 &77.25 &0.223 &0.221\T\B \\ 
RLLDN-LC\cite{paek2022row}              & LiDAR &71.73 &85.84 &78.15 &- &0.198\T\B \\ 
LiLaDet (ours)        & LiDAR &72.82 &\textbf{87.32} &\textbf{79.41} &\textbf{0.173} &\textbf{0.172} \T\B \\
\bottomrule
\end{tabular}%
}
\label{table:4}
\vspace{-1ex}
\end{table}

\subsection{Ablation Studies}
We validate the effectiveness of individual components of our proposed model and conduct the ablation experiments on the \textit{LiSV-3DLane} test set. The results are shown in Table IV where $\mathbf{BP}$ stands for the \textit{BEV Pathway} and $\mathbf{SP}$ stands for \textit{Spatial Pathway}. In Case (1), two pathway features are aggregated by element-wise addition and then forwarded to the prediction heads. The lane detection performance of our model is gradually improved by adding $\mathbf{SFWA}$ and $\mathbf{BVAT}$. The introduction of exploring spatial features in the LiDAR point cloud brings 1.83\% and 26.5\% improvements in detection performance and spatial similarity, respectively, which proves the benefits of utilizing LiDAR point information for 3D lane detection.

\begin{table}[h]
\centering
\caption{Ablation study of \textit{LiLaDet} on our \textit{LiSV-3DLane} dataset.}
\refstepcounter{table}
\resizebox{0.7\columnwidth}{!}{%
\begin{tblr}{
  width = 0.3\columnwidth,
  cells = {c},
  vline{2,5} = {-}{},
  hline{1,5} = {-}{0.08em},
  hline{2} = {-}{},
}
Case &$\mathbf{BP+SP}$           &$\mathbf{SFWA}$        &$\mathbf{BVAT}$    &$F1(\%)\uparrow$   &$CD_{3D}(m)\downarrow$ \\
(1)    &$\checkmark$     &              &              &70.52  &0.204     \\
(2)    &$\checkmark$     &$\checkmark$  &              &71.76  &0.187     \\
(3)    &$\checkmark$     &$\checkmark$  &$\checkmark$  &72.35  &0.150         
\end{tblr}
\label{table:ablation_study}
}
\vspace{-0.5em}
\end{table}

\section{CONCLUSION}
In this paper, we present a LiDAR-based surround-view 3D Lane detection dataset, \textit{LiSV-3DLane}. To handle the sparsity of manual lane annotation, we introduce an automated lane annotation pipeline to improve the labeling quality for dense prediction tasks. Subsequently, we propose a novel LiDAR-based 3D lane detection model, \textit{LiLaDet}, utilizing the spatial structural information of the LiDAR points for extracting lane markings in the point cloud scan. Extensive experiments and ablation studies prove the effectiveness of our model. For future work, we will exploit the multi-modality fusion technique of incorporating more semantic cues into the proposed LiDAR-based framework, to reduce the computational requirements of processing 3D point clouds.

\newpage
\bibliographystyle{IEEEtran}
\bibliography{main.bib}

\end{document}